\documentclass[sigconf, screen]{acmart}
\renewcommand\footnotetextcopyrightpermission[1]{} 
\AtBeginDocument{%
  }

\setcopyright{acmlicensed}
\copyrightyear{2026}
\acmYear{2026}
\acmDOI{XXXXXXX.XXXXXXX}
\acmConference[Conference acronym 'XX]{Make sure to enter the correct
  conference title from your rights confirmation email}{June 03--05,
  2018}{Woodstock, NY}
\acmISBN{978-1-4503-XXXX-X/2018/06}

\usepackage[table]{xcolor}
\usepackage{booktabs}
\usepackage{tabularx}
\usepackage{graphicx}
\usepackage{subcaption}
\usepackage{array}
\usepackage{pifont}
\usepackage{float}      
\usepackage{placeins}

\newcolumntype{Y}{>{\centering\arraybackslash}X}
\definecolor{mycustompink}{RGB}{255, 228, 225}
\begin{document}


\title{GTPBD-MM: A Global Terraced Parcel and Boundary Dataset with Multi-Modality}

\author{
\textbf{Zhiwei Zhang}$^{1,*}$,
\textbf{Xingyuan Zeng}$^{1,*}$,
\textbf{Xinkai Kong}$^{1,*}$,
\textbf{Kunquan Zhang}$^{1}$,
\textbf{Haoyuan Liang}$^{1}$,
\textbf{Bohan Shi}$^{5}$,
\textbf{Juepeng Zheng}$^{1,6,\dagger}$ 
\textbf{Jianxi Huang}$^{3,4}$,
\textbf{Yutong Lu}$^{1,6}$,
\textbf{Haohuan Fu}$^{2,6}$ \\
{
$^{1}$Sun Yat-sen University, 
$^{2}$Tsinghua Shenzhen International Graduate School \\
$^{3}$China Agricultural University, 
$^{4}$Southwest Jiaotong University \\
$^{5}$Northeastern University, 
$^{6}$National Supercomputing Center in Shenzhen \\
$^{*}$Equal contribution, $^{\dagger}$Corresponding author. 
}
}
\renewcommand{\shortauthors}{Trovato et al.}

\begin{abstract}

Agricultural parcel extraction plays an important role in remote sensing-based agricultural monitoring, supporting parcel surveying, precision management, and ecological assessment. However, existing public benchmarks mainly focus on regular and relatively flat farmland scenes. In contrast, terraced parcels in mountainous regions exhibit stepped terrain, pronounced elevation variation, irregular boundaries, and strong cross-regional heterogeneity, making parcel extraction a more challenging problem that jointly requires visual recognition, semantic discrimination, and terrain-aware geometric understanding. Although recent studies have advanced visual parcel benchmarks and image-text farmland understanding, a unified benchmark for complex terraced parcel extraction under aligned image-text-DEM settings remains absent. To fill this gap, we present \textbf{GTPBD-MM}, the first multimodal benchmark for global terraced parcel extraction. Built upon GTPBD, GTPBD-MM integrates high-resolution optical imagery, structured text descriptions, and DEM data, and supports systematic evaluation under \textit{Image-only}, \textit{Image+Text}, and \textit{Image+Text+DEM} settings. We further propose \textbf{E}levation and \textbf{T}ext guided \textbf{Terra}ced parcel network (\textbf{ETTerra}), a multimodal baseline for terraced parcel delineation. Extensive experiments demonstrate that textual semantics and terrain geometry provide complementary cues beyond visual appearance alone, yielding more accurate, coherent, and structurally consistent delineation results in complex terraced scenes. 
Dataset and supplementary materials are publicly available at \url{https://github.com/Z-ZW-WXQ/GTPBD-MM}.

\end{abstract}






\maketitle

\section{Introduction}
\begin{figure}
    \centering
    \includegraphics[width=1\linewidth]{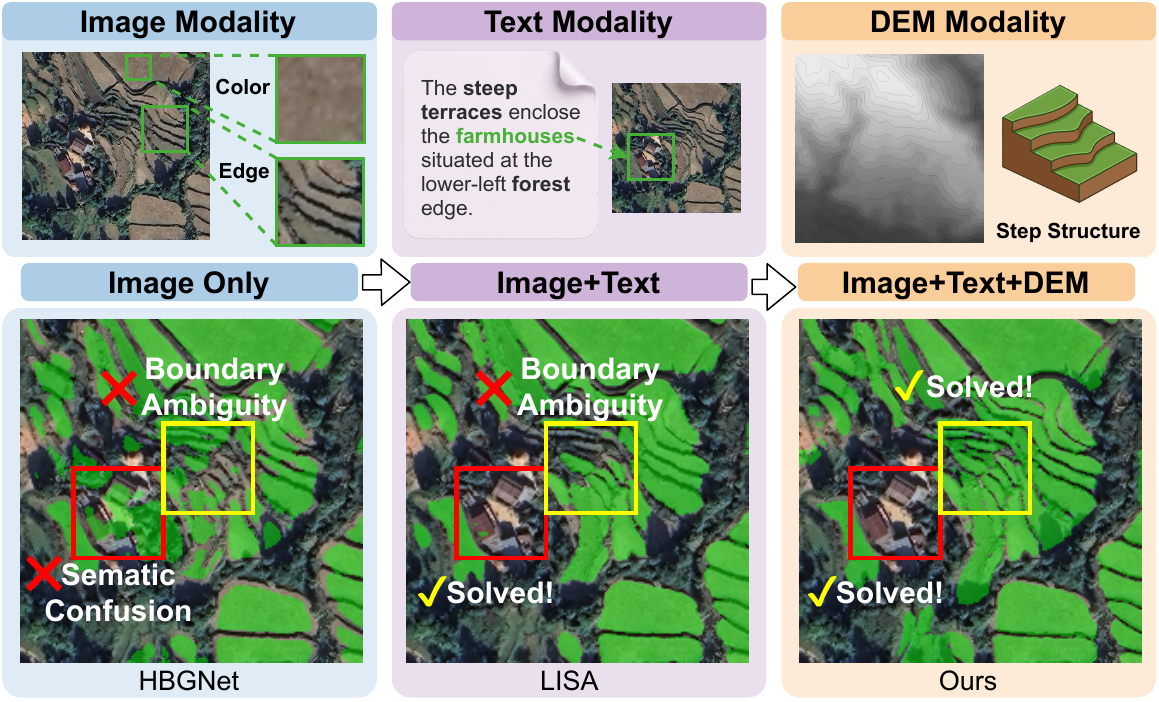}
    \vspace{-1.8em}

    \caption{Overview of terraced parcel extraction challenges and multimodal motivation. The top row illustrates the complementary information provided by image, text, and DEM modalities. 
    The bottom row compares three input settings: the Image-only model (HBGNet~\cite{FHAPD/HBGNet}) suffers from semantic confusion (red box) and boundary ambiguity (yellow box); the Image+Text model (LISA~\cite{lai2024lisa}) mitigates semantic confusion but still fails to resolve the boundary ambiguity; by jointly modeling image, text, and DEM, our method effectively addresses both issues and yields more complete and structurally consistent parcel delineation results.}
    
    \vspace{-2.0em}
    \label{fig:overview}
\end{figure}

\begin{table*}[t]
\centering
\scriptsize
\renewcommand{\arraystretch}{1.08}
\caption{Comparison of representative datasets for agricultural parcel analysis in terms of research focus, input modality, spatial resolution, annotation type, geographic coverage, and whether terraced scenes are included.}
\vspace{-1.5em}
\label{tab:dataset_compare}
\begin{tabularx}{\textwidth}{l l l l l l l c}
\toprule
\textbf{Dataset} & \textbf{Research Focus} & \textbf{Pub.} & \textbf{Input} & \textbf{Resolution} & \textbf{Annotation} & \textbf{Coverage} & \textbf{Terraced Scenes} \\
\midrule
GFSAD30~\cite{GFSAD30}      & Cropland Mapping                     & USGS'21      & Image             & 30\,m       & Mask                 & Global         & No  \\
GTM~\cite{li202510}         & Cropland Mapping                     & JAG'25       & Image             & 10\,m       & Mask                 & Global   & Yes \\
AI4Boundaries~\cite{AI4Boundaries} & Parcel Delineation           & ESSD'23      & Image             & 10\,m, 1\,m & Boundary, Vector     & Europe         & No  \\
FHAPD~\cite{FHAPD/HBGNet}   & Parcel Delineation                   & ISPRS'25     & Image             & 1--2\,m     & Mask                 & China          & No  \\
FTW~\cite{FTW}              & Parcel Delineation                   & AAAI'25      & Image             & 10\,m       & Mask, Parcel         & Global         & No  \\
GTPBD~\cite{zhang2025gtpbd} & Parcel Delineation                   & NeurIPS'25   & Image             & 0.5--0.7\,m & Boundary, Mask, Parcel & Global & Yes \\
FarmSeg-VL~\cite{farmsegvl} & Image-Text Farmland Understanding    & ESSD'25      & Image + Text      & 0.5--2\,m   & Mask, Caption        & China          & No  \\
\bottomrule
\rowcolor{pink!30}
\textbf{GTPBD-MM (Ours)} & \textbf{Multimodal Terraced Parcel Delineation} & \textbf{This work} & \textbf{Image + Text + DEM} & \textbf{0.5--0.7\,m} & \textbf{Boundary, Mask, Parcel, Caption} & \textbf{Global} & \textbf{Yes} \\
\bottomrule
\end{tabularx}
\vspace{-1.5em}
\end{table*}

Agricultural parcel extraction is a fundamental task in remote sensing-based agricultural monitoring, with broad importance for parcel surveying, precision management, and ecological assessment~\cite{spencer1961origin, weiss2020remote}.  However, although recent deep learning methods have substantially improved farmland segmentation performance~\cite{11396383}, current public benchmarks are still mainly designed for regular and relatively flat conventional farmland scenes~\cite{wang2023survey, hadir2025comparative}, leaving the terraced scenario insufficiently explored. Actually, terraced parcels are a widespread agricultural landscape in mountainous and hilly regions worldwide, especially in developing countries~\cite{tarolli2018terraced, modica2017abandonment, li202510}. 
In contrast to conventional farmland, terraced parcels are shaped not only by image appearance, but also by elevation variation, slope transitions, and step-like terrain structures. 
As a result, terraced parcel extraction is not simply a harder version of conventional farmland segmentation, but a more complex parsing problem that jointly requires accurate boundary recognition~\cite{wang2006boundary}, reliable target discrimination, and structured terrain understanding~\cite{zhang2025gtpbd}.

As illustrated in Fig.~\ref{fig:overview}, terraced parcel extraction is mainly challenged by two key difficulties: \textbf{semantic confusion} and \textbf{boundary ambiguity}. Semantic confusion occurs when visually similar non-target objects, such as houses, ponds, roads, ridges, and bare land, are mistaken for terraced parcels, making target discrimination more difficult. Boundary ambiguity arises when adjacent terrace units share similar appearance, while their true boundaries are determined by elevation discontinuities and step-like terrain structures, often leading to incomplete extraction, blurred boundaries, and erroneous merging across neighboring terrace steps.

Most existing parcel extraction methods are still built in an image-only manner~\cite{reaunet,FHAPD/HBGNet,SEANet,xie2026cnn,li2024comprehensive,wu2023cmtfnet}. Existing image-only methods cannot effectively resolve these two issues. Although the image modality provides direct appearance cues such as color, texture, shape, and local edges, which are useful for region localization and coarse boundary delineation~\cite{11224516,10589602}, it remains insufficient in complex terraced scenes. As shown in Fig.~\ref{fig:overview}, the image-only model (\textit{e.g.}, {HBGNet}~\cite{FHAPD/HBGNet}) is prone to both semantic confusion and boundary ambiguity. Introducing text modality can alleviate the first problem by providing semantic priors about category attributes, scene composition, and spatial relationships. 
Recently, several studies have explored parcel extraction by jointly leveraging image and text modalities~\cite{huang2021text,zhang2023text2seg,chen2023generative,luddecke2022image,lauriola2022introduction,fsvlm/fit,farmsegvl}. These methods can effectively alleviate semantic confusion by introducing semantic priors about target categories, scene composition, and spatial relationships. In particular, the image+text model (\textit{e.g.}, {LISA}~\cite{lai2024lisa}) can better distinguish true parcels from visually similar non-target regions, thereby suppressing typical semantic confusion. However, since text does not provide explicit terrain geometry, such methods still cannot fundamentally resolve the boundary ambiguity caused by missing elevation structure. Moreover, existing studies are still largely centered on regular and relatively flat farmland scenes, while complex terraced parcels remain rarely investigated.

This limitation is particularly pronounced in terraced environments, where parcel layouts are strongly shaped by mountainous relief. In fact, the geometric structure of terraced parcels is naturally aligned with the underlying terrain, as terrace boundaries are often formed along elevation discontinuities, slope transitions, and step-like landforms. This inherent consistency makes Digital Elevation Model (DEM) a particularly suitable source of structural information for terraced parcel extraction. By providing terrain-aware geometric cues~\cite{tadono2014precise,spano2018gis,ma2024multilevel}, 
DEM can help recover structurally consistent parcel boundaries in visually cluttered scenes and reduce under-segmentation, boundary blurring, and terrace-step merging~\cite{ma2025unified,cao2019bundle,colwell2000mid,pike1988geometric}. As shown in Fig.~\ref{fig:overview}, the progression from the image-only model (\textit{e.g.}, {HBGNet}~\cite{FHAPD/HBGNet}) to the image-text model (\textit{e.g.}, {LISA}~\cite{lai2024lisa}) and finally to our image-text-DEM model demonstrates that image, text, and DEM play complementary roles in terraced parcel extraction, and that only their joint modeling can reliably address both semantic confusion and boundary ambiguity.


Motivated by the above challenges, we argue that complex terraced parcel extraction should be studied as a new multimodal problem requiring the joint modeling of {appearance}, {semantics}, and {geometry}. However, existing public benchmarks still lack a unified multimodal infrastructure that can align image, text, and terrain geometry in complex terraced scenes. Although {GTPBD} \cite{zhang2025gtpbd} has already laid a strong foundation with global terraced scenes, fine-grained annotations, and multi-task benchmarking, it does not yet support systematic multimodal research under aligned image-text-DEM settings, so that causing existing appraoches faces above two challenges (\textit{i.e.}, semantic confusion and boundary ambiguity) in terraced parcel mapping and boundary delineation. 
To fill this gap, we propose \textbf{GTPBD-MM}, a multimodal benchmark for complex terraced parcel extraction built upon GTPBD, which unifies three complementary modalities: high-resolution optical imagery, DEM, and text descriptions. Based on this dataset, we further propose a multimodal baseline, \textbf{E}levation-\textbf{T}ext guided \textbf{Terra}ced parcel network (\textbf{ETTerra}), to validate the collaborative effect of image appearance, textual semantics, and terrain geometry in complex terraced scene parsing, which effectively address semantic confusion and boundary ambiguity in terraced parcel extraction. 

Overall, our contributions are summarized as follows:

$\bullet$ We propose \textbf{GTPBD-MM}, the first multimodal benchmark for global complex terraced parcel extraction that jointly aligns high-resolution imagery, text descriptions, and DEM information, providing a new public foundation for multimodal terraced scene understanding.

$\bullet$ We propose \textbf{ETTerra}, a multimodal baseline that integrates visual appearance, textual semantics, and terrain geometry, and serves as a benchmark model for studying collaborative tri-modal parsing in complex terraced environments.

$\bullet$ We establish extensive benchmark evaluations on \textbf{GTPBD-MM}, covering different modality settings (including {Image-only}, {Image+Text}, and {Image+Text+DEM}) and multiple evaluation levels (including {pixel-}, {boundary-}, and {object-level} metrics), enabling systematic analysis of multimodal terraced parcel extraction.
\vspace{-1.0em}

\begin{figure*}[t]
    \centering
    \includegraphics[width=\textwidth]{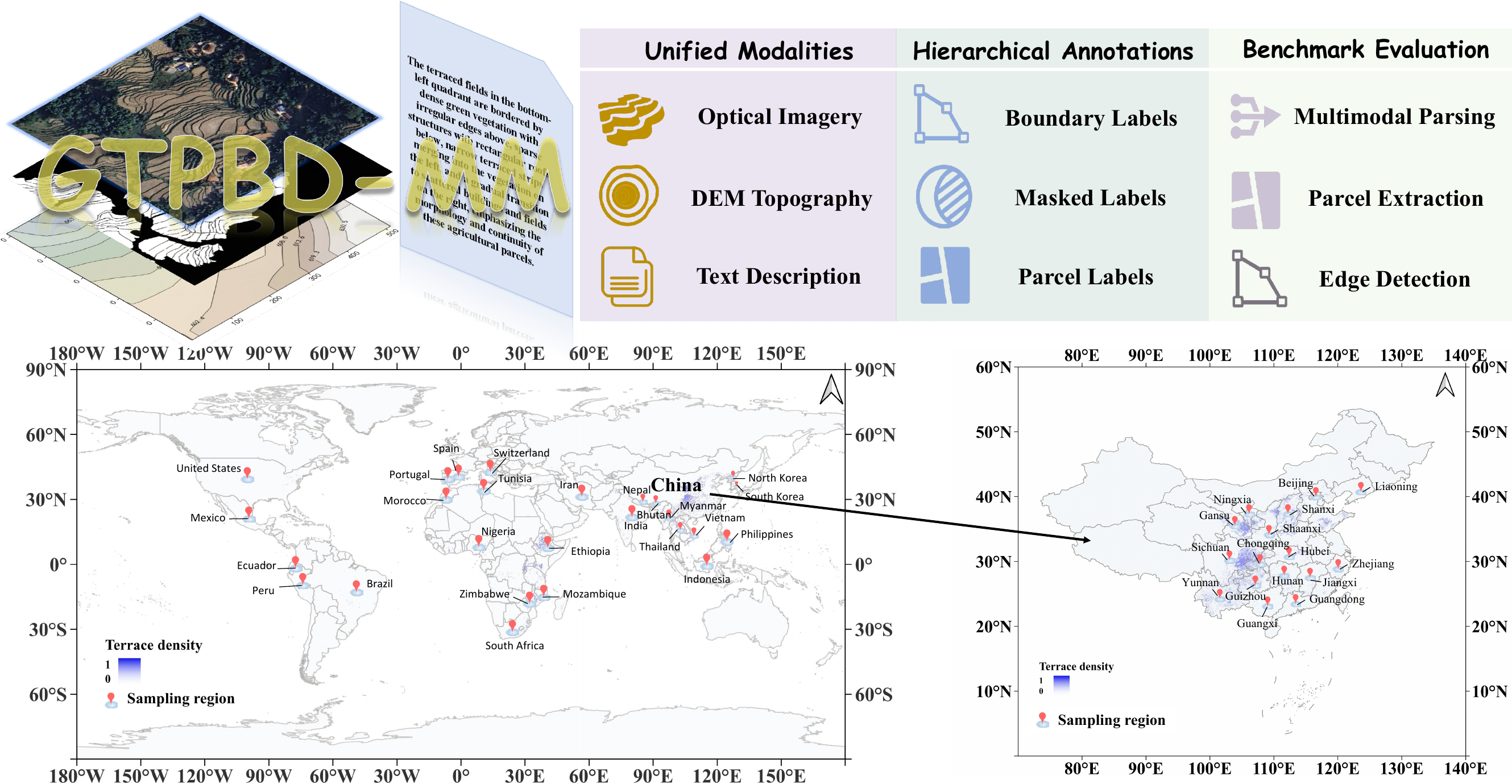}
    \vspace{-2em}
    \caption{Overview of GTPBD-MM. Top: unified multimodal design with aligned modalities, hierarchical annotations, and evaluation tasks. Bottom: spatial sampling distribution, with global coverage and a zoom-in view of China.}
    \label{fig:dataset_overview}
    \vspace{-0.5em}
\end{figure*}

\section{Related Work}
\subsection{Agricultural Parcel Benchmarks}

As summarized in Table~\ref{tab:dataset_compare}, existing datasets for agricultural parcel analysis can be broadly grouped into three categories: cropland mapping, parcel delineation, and image-text farmland understanding. Datasets in the first two categories, such as GFSAD30~\cite{GFSAD30}, GTM~\cite{li202510}, AI4Boundaries~\cite{AI4Boundaries}, FHAPD~\cite{FHAPD/HBGNet}, and FTW~\cite{FTW}, have progressively improved spatial resolution, annotation quality, and geographic coverage, thereby providing a solid foundation for agricultural parcel analysis. Among them, GTPBD~\cite{zhang2025gtpbd} further advances parcel delineation benchmarks by shifting the focus from regular farmland to globally distributed complex terraced scenes.
A more recent line of work introduces language into farmland understanding. FSVLM~\cite{fsvlm/fit} and FarmSeg-VL~\cite{farmsegvl} show that text can provide useful semantic descriptions beyond visual appearance alone. However, these datasets are still mainly designed for regular farmland scenes. For terraced parcels, whose structures are closely coupled with terrain relief, image-text organization alone remains insufficient, making DEM a particularly important modality.

However, especially for terraced parcel extraction that are strongly shaped by mountainous relief, a benchmark that jointly organizes {image}, {text}, and {DEM} for complex terraced parcel extraction is still missing. Therefore, our \textbf{GTPBD-MM} is designed to fill this gap.
\subsection{Parcel Extraction and Multimodal Modeling}

Existing methods for agricultural parcel extraction mainly follow two directions. The first focuses on image-only parcel delineation, including both general segmentation models~\cite{unet,pspnet,chen2017rethinking,segformer,mask2former} and parcel-oriented methods such as CMTFNet~\cite{wu2023cmtfnet}, REAUNet~\cite{reaunet}, HBGNet~\cite{FHAPD/HBGNet} and SLFNet~\cite{tong2026slfnet}. These methods improve parcel extraction through stronger boundary modeling and edge enhancement, but they still rely primarily on two-dimensional visual appearance. As a result, they remain limited in addressing the two key challenges of terraced parcel extraction, namely {semantic confusion} and {boundary ambiguity} (See Fig.~\ref{fig:overview}). 
The second direction introduces language into farmland segmentation and scene understanding. Methods such as FSVLM~\cite{fsvlm/fit} and FarmSeg-VLM~\cite{wu2025farmseg_vlm}, together with language-conditioned paradigms such as referring and reasoning segmentation~\cite{chen2025rsrefseg,yuan2024rrsis,dong2025diffris,ding2021vision,lai2024lisa,wang2024llm,shen2025reasoning,ren2024pixellm}, show that text can enhance semantic discrimination and target localization. This is helpful for alleviating semantic confusion, but still insufficient for resolving the boundary ambiguity in terraced scenes, since explicit terrain geometry is absent.

Overall, existing methods have covered appearance-based parcel delineation and appearance-semantics-based farmland understanding, yet unified modeling of {appearance}, {semantics}, and {terrain geometry} remains lacking for complex terraced parcel extraction. Our goal is therefore to jointly utilize image, text, and DEM, and yields more structurally consistent parcel delineation.

\begin{figure*}[!t]
    \centering
    \includegraphics[width=\textwidth]{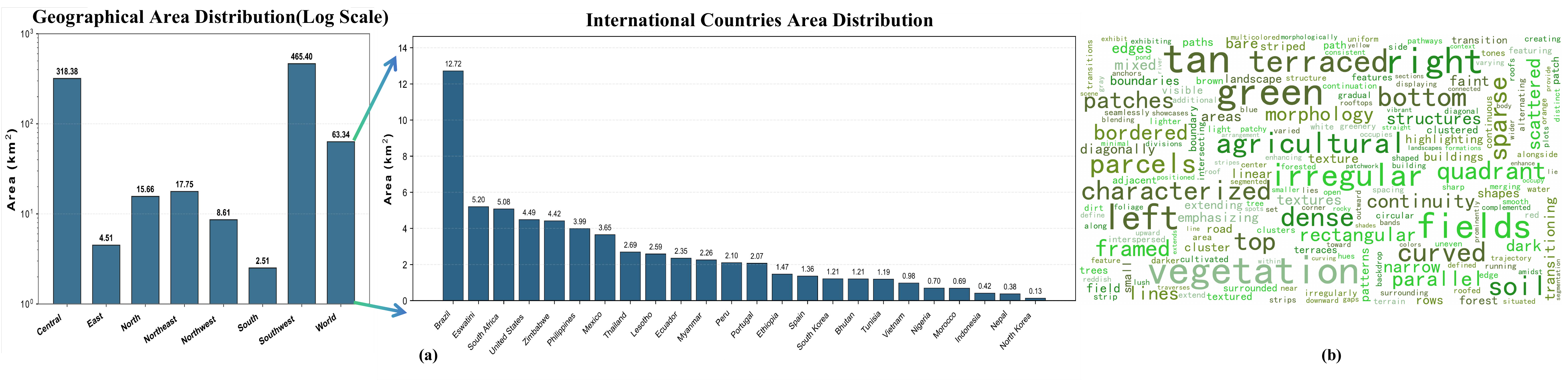}
    \vspace{-2em}
    \caption{Dataset statistics of GTPBD-MM. (a) Regional- and Country-level area distribution. (b) Word cloud of text descriptions.}
    \label{fig:stats}
    \vspace{-1em}
\end{figure*}

\section{GTPBD-MM Dataset}
\subsection{Dataset Overview}

GTPBD-MM is a multimodal benchmark built upon GTPBD for complex terraced parcel understanding.
Each sample consists of a spatially aligned high-resolution optical image, DEM, task-oriented text description, and three-level annotations, including mask, boundary, and parcel labels. By unifying visual appearance, scene semantics, and terrain geometry within the same sample, GTPBD-MM provides a common data foundation for multimodal parcel parsing in complex terraced scenes.

The dataset is sampled from globally distributed terraced regions, covering more than 900\,km$^2$ across 25 countries worldwide while maintaining systematic coverage over the seven geographical divisions of China.
Building upon GTPBD, we further extend the dataset with samples from 11 additional countries(e.g., Nepal, Indonesia, and Zimbabwe), leading to broader spatial coverage and richer diversity across geomorphological backgrounds and regional styles. With its unified sample organization and annotation scheme, GTPBD-MM supports multimodal parsing, parcel extraction, and edge detection under a relatively complete benchmark setting.
More details and examples of the dataset are provided in Appendix A.

\subsection{Dataset Construction and Statistics}

For dataset construction, GTPBD-MM inherits the fine-grained annotation system of GTPBD and further augments each sample with DEM and text modalities. The high-resolution optical imagery is mainly sourced from GaoFen-2. For DEM, we acquire elevation data corresponding to the spatial extent of each optical image, and apply resampling, cropping, and registration to ensure strict spatial alignment with the optical image and the three-level annotations. For the text modality, we construct structured descriptions tailored to terraced parcel extraction, focusing on scene-level layout, local parcel morphology, and surrounding spatial relations, thereby providing task-relevant semantic information beyond generic captions.

Figure~\ref{fig:stats} presents an overview of GTPBD-MM from the perspectives of area statistics and textual characteristics. As shown in Fig.~\ref{fig:stats}(a), the dataset covers a wide range of geographical regions and countries in terms of spatial area. On the one hand, typical terraced clusters in Southwest China remain the major coverage regions; on the other hand, the international samples further broaden the dataset coverage across different countries and geomorphological settings, making the benchmark both regionally representative and geographically diverse. Figure~\ref{fig:stats}(b) shows the high-frequency words in the text descriptions, such as {terraced}, {irregular}, {curved}, {dense}, and {vegetation}. These words mainly characterize terrace morphology, boundary structure, land cover, and spatial relations, indicating that the text modality provides semantic priors complementary to image appearance and DEM geometry.

Building on this construction pipeline, 
GTPBD-MM naturally establishes a unified multimodal benchmark setting. This allows systematic comparisons under Image-only, Image+Text, and Image+Text+DEM settings, and provides a standardized data foundation for appearance, semantics and geometry collaborative modeling in complex terraced scenes.

\begin{table*}[t]
\centering
\caption{
Comprehensive benchmark results on \textbf{GTPBD-MM}. 
\textit{Gen. Sem. Seg.}, \textit{Parcel Delin.}, \textit{VL Seg.}, and \textit{MM Parcel Delin.} denote \textit{General Semantic Segmentation}, \textit{Parcel Delineation}, \textit{Reasoning Segmentation}, and \textit{Multimodal Parcel Delineation}, respectively. 
$I$, $T$, and $D$ denote image, text, and DEM inputs, respectively. 
\textit{Pub.} indicates the publication venue/year. 
Best and second-best completed results are highlighted in \textbf{bold} and \underline{underline}, respectively.
}
\label{tab:gtpbdmm_full_compact}
\scriptsize
\setlength{\tabcolsep}{3.6pt}
\renewcommand{\arraystretch}{1.03}
\resizebox{\textwidth}{!}{
\begin{tabular}{llllccccc cc ccc}
\toprule
 &  &  & 
& \multicolumn{5}{c}{\textbf{Pixel-level}}
& \multicolumn{2}{c}{\textbf{Edge-level}}
& \multicolumn{3}{c}{\textbf{Object-level}} \\
\cmidrule(lr){5-9} \cmidrule(lr){10-11} \cmidrule(lr){12-14}
\textbf{Model} & \textbf{Method Family} & \textbf{Input} & \textbf{Pub.} 
& \textbf{Rec.$\uparrow$} & \textbf{F1$\uparrow$} & \textbf{OA$\uparrow$} & \textbf{mIoU$\uparrow$} & \textbf{mAcc$\uparrow$}
& \textbf{OIS$\uparrow$} & \textbf{ODS$\uparrow$}
& \textbf{GOC$\downarrow$} & \textbf{GUC$\downarrow$} & \textbf{GTC$\downarrow$} \\
\midrule
U-Net~\cite{unet}         & Gen. Sem. Seg.    & I     & MICCAI'15  & 57.15 & 65.42 & 72.67 & 55.87 & 71.32 & 22.97 & 15.57 & 44.51 & 39.47 & 47.11 \\
PSPNet~\cite{pspnet}        & Gen. Sem. Seg.    & I     & CVPR'17    & 63.12 & 69.39 & 72.83 & 58.29 & 72.97 & 18.93 & 12.71 & \textbf{20.00} & 46.47 & 40.63 \\
Deeplabv3~\cite{chen2017rethinking} & Gen. Sem. Seg.    & I     & CVPR'18  & 81.07 & 77.77 & 79.03 & 65.25 & 79.21 & 28.29 & 20.04 & 21.76 & 40.85 & \underline{38.42} \\
SegFormer~\cite{segformer}     & Gen. Sem. Seg.    & I     & NeurIPS'21 & 69.07 & 71.66 & \underline{79.14} & 63.75 & 77.21 & 22.97 & 15.70 & 26.10 & 41.45 & 41.10 \\
Mask2Former~\cite{mask2former}   & Gen. Sem. Seg.    & I     & CVPR'22    & 64.20 & 69.69 & 74.73 & 58.93 & 73.82 & 36.00 & 26.33 & 33.92 & 38.03 & 41.98 \\
REAUNet~\cite{reaunet}       & Parcel Delin.     & I     & JAG'24     & \underline{84.61} & 77.02 & 77.15 & 62.80 & 77.80 & 36.35 & 27.42 & 28.84 & 39.67 & 40.82 \\
HBGNet~\cite{FHAPD/HBGNet}        & Parcel Delin.     & I     & ISPRS'25   & 58.14 & 64.42 & 70.95 & 54.05 & 69.84 & \underline{57.62} & \underline{42.64} & 81.60 & \textbf{31.98} & 66.93 \\
LaSagnA~\cite{wei2024lasagna}     & Reasoning Seg. & I+T   & Arxiv'24    & 74.41 & 75.31 & 77.96 & 63.60 & 77.64 & 44.57 & 33.09 & 35.29 & 36.07 & 42.06 \\
LISA ~\cite{lai2024lisa}         & Reasoning Seg.    & I+T   & CVPR'24    & 80.65 & 77.67 & 79.02 & 65.23 & 79.16 & 42.06 & 31.10 & 28.27 & 38.77 & 39.94 \\
PixelLM ~\cite{ren2024pixellm}         & Reasoning Seg.    & I+T   & CVPR'24    & 84.34 & \underline{78.53} & 79.13 & \underline{65.45} & \underline{79.59} & 29.62 & 20.33 & \underline{21.52} & 43.92 & 39.56 \\
FSVLM~\cite{fsvlm/fit}         & MM Parcel Delin.  & I+T   & TGRS'25    & 74.09 & 76.10 & 78.95 & 64.88 & 78.53 & 43.34 & 31.87 & 35.20 & 38.59 & 44.02 \\
ETTerra (Ours) & MM Parcel Delin.  & I+T+D & This work  & \textbf{86.40} & \textbf{80.80} & \textbf{81.51} & \textbf{68.73} & \textbf{81.76} & \textbf{63.72} & \textbf{49.52} & 22.40 & \underline{35.64} & \textbf{36.78} \\
\bottomrule
\end{tabular}}
\vspace{-2em}
\end{table*}



\begin{figure}
    \centering
    \includegraphics[width=1\linewidth]{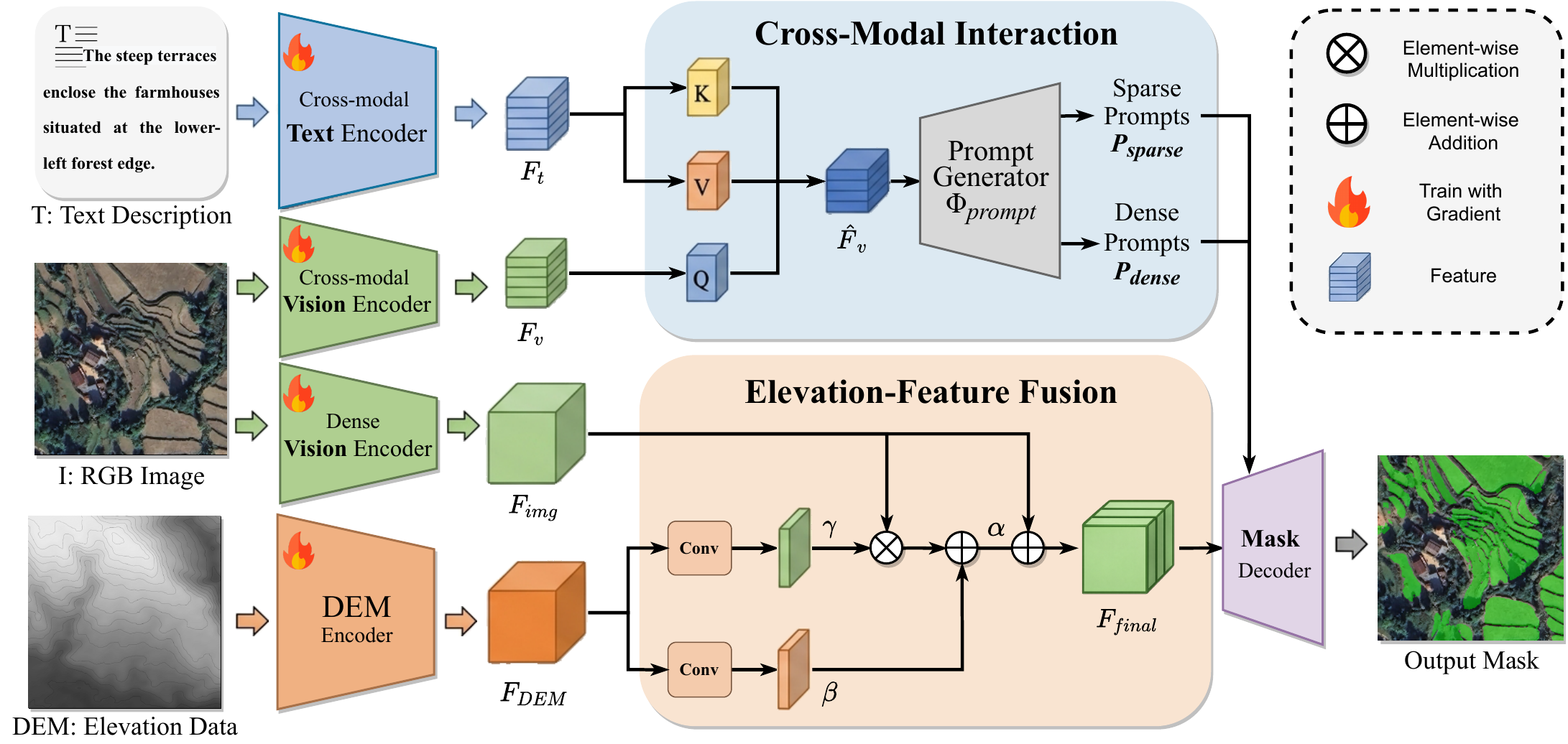}
    \caption{Overview of our proposed Elevation and Text Guided Terraced Parcel Network (ETTerra), which integrates cross-modal interaction and spatial terrain modulation for multimodal terraced parcel extraction.}
    \label{fig:method}
\end{figure}

\section{Proposed Method: ETTerra}
To address the challenges of \textbf{semantic confusion} and \textbf{boundary ambiguity} in terrace scenes, this paper proposes a multimodal decoupled segmentation architecture, named \textbf{E}levation-\textbf{T}ext guided \textbf{Terra}ced parcel network (\textbf{ETTerra}). This architecture decouples terrace extraction into two parallel branches: a {cross-modal semantic enhancement branch} that utilizes textual semantic priors to alleviate semantic confusion, and an {elevation-guided boundary reinforcement branch} that leverages elevation geometric priors to mitigate boundary ambiguity.
Full hyperparameter configurations and hardware specifications are provided in Appendix B.1. 
\subsection{Cross-Modal Semantic Enhancement}

In Image-only segmentation, visually similar non-target regions (\textit{e.g.}, houses, bare land) are prone to causing misclassification. This branch utilizes the category and scene priors embedded in the text description $T$ to guide the model in target discrimination.

Given an RGB image $I$ and a text description $T$, visual features $F_{v} \in \mathbb{R}^{D \times N_{v}}$ and text features $F_{t} \in \mathbb{R}^{D \times N_{t}}$ are first extracted through a pre-trained {cross-modal vision encoder} and a {cross-modal text encoder}, respectively, and projected into a shared latent space. To enable the visual features to perceive the macroscopic scene described by the text and thereby suppress background interference, this branch performs {cross-modal interaction}, conditioning on $F_{t}$ to semantically enhance $F_{v}$ via Multi-Head Cross-Attention (MHCA):

\begin{equation}
\hat{F}_{v} = \text{MHCA}(F_{v}, F_{t}, F_{t}) + F_{v}
\end{equation}

Here, $F_{v}$ serves as the Query (Q), while $F_{t}$ serves as the Key (K) and Value (V). Through cross-modal interaction, the network actively suppresses the feature responses of visually similar backgrounds by exploiting textual priors. Subsequently, the aligned cross-modal features $\hat{F}_{v}$ are fed into a prompt generator $\Phi_{prompt}$, where they are transformed into dense prompts ($P_{dense}$) and sparse prompts ($P_{sparse}$) for the mask decoder, providing explicit semantic constraints for the segmentation process.

\subsection{Elevation Prior-Guided Boundary Reinforcement}

The similar appearance of adjacent terrace units easily causes boundary ambiguity, whereas the actual physical boundaries of terraces typically manifest as abrupt elevation changes and stepped terrain. Therefore, this branch introduces DEM data to recover structurally consistent terrace boundaries.

The RGB image is first processed by a {dense vision encoder} to extract spatial features $F_{img} \in \mathbb{R}^{C \times H \times W}$ containing local textures. Meanwhile, the DEM data is independently encoded into elevation features $F_{DEM} \in \mathbb{R}^{C \times H \times W}$ by a lightweight fully convolutional network acting as the {DEM encoder}. This independent encoding strategy avoids information loss caused by the early fusion of heterogeneous data. To modulate the visual features using topographical geometric structures, we design a {Elevation-Feature Fusion} module: it generates spatially adaptive scaling factors $\gamma$ and shifting factors $\beta$ using convolutional layers based on $F_{DEM}$, and applies an element-wise affine transformation to $F_{img}$. To ensure training stability and preserve local image details, the modulated features are aggregated with the original visual features via a zero-initialized residual connection:

\begin{equation}
F_{final} = F_{img} + \alpha \cdot (\gamma \odot F_{img} + \beta)
\end{equation}

where $\odot$ denotes the Hadamard product ($\otimes$ in the diagram), and $\alpha$ is a learnable scalar initialized to $0$. This fusion mechanism explicitly utilizes the slope drop cues derived from the DEM, dynamically sharpening the blurred terrace boundaries at the feature level.

\subsection{Dual-Branch Collaborative Mask Decoding}

The high-resolution visual features $F_{final}$, which are fused with topographical priors, along with the prompt information generated by the semantic enhancement branch, are jointly fed into the {mask decoder}. Guided simultaneously by the {semantic constraints} provided by the text and the {structural features} reinforced by the DEM, the decoder performs pixel-level decoding, ultimately generating the complete and boundary-distinct terrace {output mask}.

\begin{figure*}[t]
    \centering
    \includegraphics[width=1\linewidth]{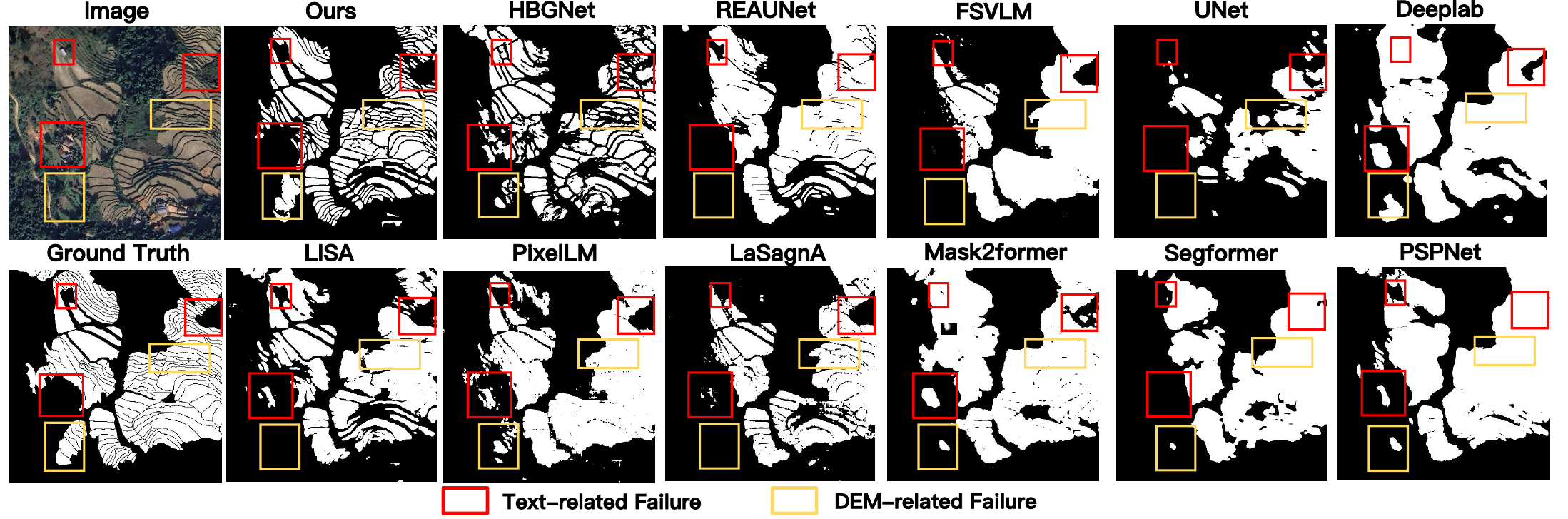}
    \vspace{-1em}
    \caption{Qualitative comparison of different methods on GTPBD-MM. Red boxes highlight typical semantic confusion in non-parcel regions without sufficient textual guidance, while yellow boxes show geometric recovery errors caused by the absence of DEM cues. Our ETTerra produces more complete, coherent, and structurally consistent parcel delineation results.}
    \label{fig:model_compare}
    \vspace{-1.8em}
\end{figure*}

\begin{figure}[t]
    \centering
    \includegraphics[width=1.0\linewidth]{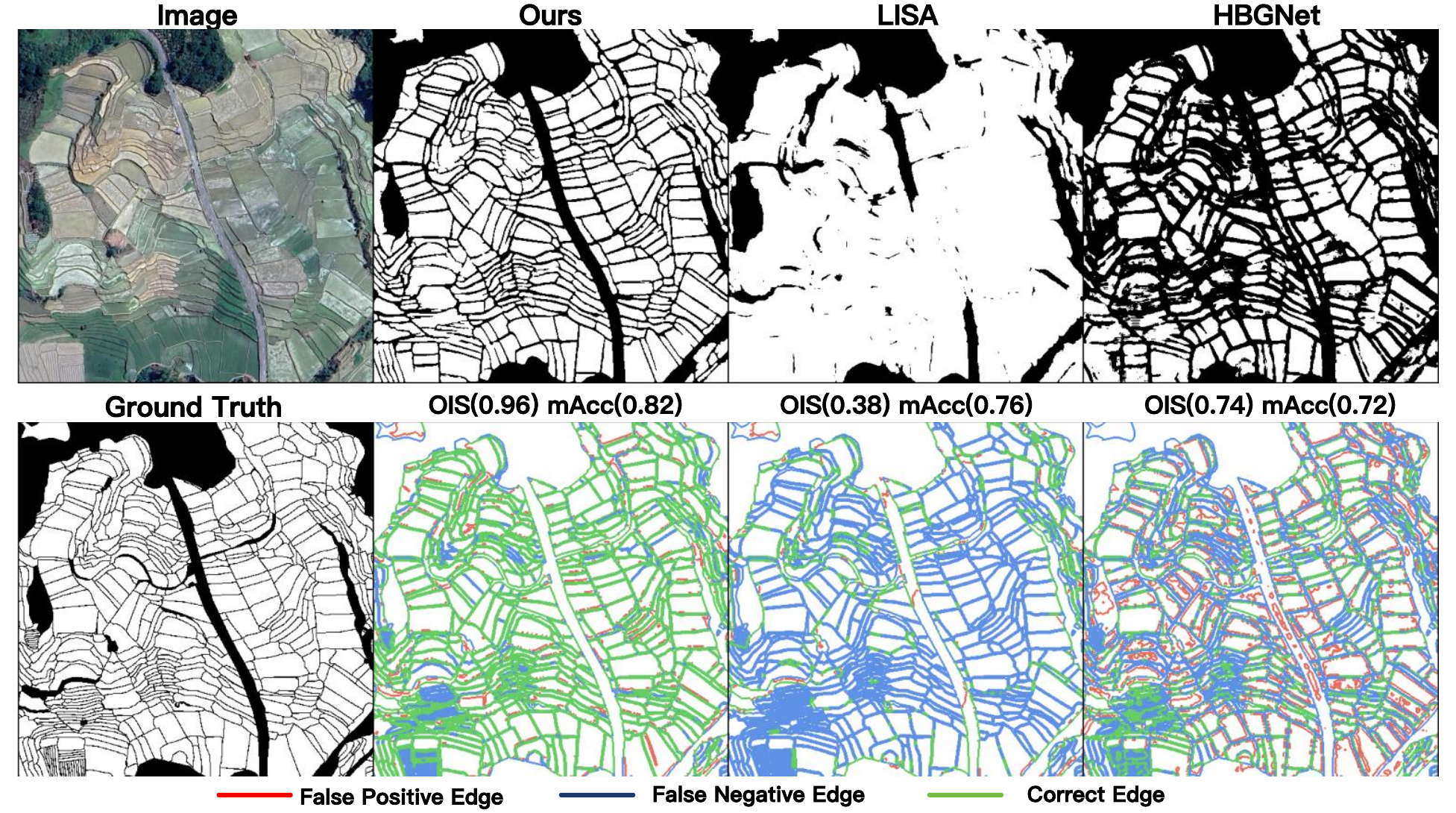}
    \caption{Edge-level error analysis of different methods on GTPBD-MM. We visualize correct edges, false positive edges, and false negative edges together with OIS and mAcc.}
    \vspace{-1.5em}
    \label{fig:edge}
\end{figure}


\section{Benchmark and Evaluation}

\subsection{Benchmark Protocol}

To systematically evaluate the benchmark value of \textbf{GTPBD-MM} for complex terraced parcel extraction, we consider three input settings under a unified data split: \textbf{Image-only}, \textbf{Image+Text}, and \textbf{Image+Text+DEM}. Compared with conventional single-modality evaluation, this design enables GTPBD-MM to assess not only the basic segmentation capability based on visual appearance, but also the complementary roles of textual semantics and terrain geometry in complex terrace understanding.

We include representative methods from multiple methodological families, including five {general semantic segmentation} methods (\textit{i.e.}, U-Net~\cite{unet}, PSPNet~\cite{pspnet}, Deeplabv3~\cite{chen2017rethinking}, SegFormer~\cite{segformer}, and Mask2Former~\cite{mask2former}), two {parcel delineation} methods (\textit{i.e.}, REAUNet~\cite{reaunet} and HBGNet~\cite{FHAPD/HBGNet}), three {reasoning segmentation} methods (\textit{i.e.}, LISA~\cite{lai2024lisa}, PixelLM~\cite{ren2024pixellm}, LaSagnA~\cite{wei2024lasagna}), and a {multimodal parcel delineation} method (\textit{i.e.}, FSVLM~\cite{fsvlm/fit}). 
Detailed descriptions for each baseline method are provided in Appendix B.2. 
Such a benchmark setting covers major technical paradigms from pure visual segmentation and language-guided segmentation to multimodal parcel modeling.

We adopt a three-level evaluation protocol. \textbf{Pixel-level} metrics include Recall, F1, OA, mIoU, and mAcc, which evaluate region-level segmentation quality. \textbf{Edge-level} metrics include OIS and ODS, which measure boundary recovery quality. \textbf{Object-level} metrics include GOC, GUC, and GTC, which reflect geometric error and structural consistency at the object level. This protocol provides a comprehensive assessment of model performance from region, boundary, and object perspectives. More details could be found in Appendix C.

\vspace{-0.4em}

\subsection{Benchmark Results and Analysis}

Table~\ref{tab:gtpbdmm_full_compact} presents the comprehensive benchmark results on GTPBD-MM. Among all completed baselines, \textbf{ETTerra} achieves the best overall performance, ranking first in \textbf{Recall, F1, OA, mIoU, mAcc, OIS, ODS}, and \textbf{GTC}, with \textbf{mIoU = 68.73}, \textbf{ODS = 49.52}, and \textbf{GTC = 36.78}. These results indicate that jointly modeling image appearance, textual semantics, and terrain geometry not only improves overall pixel-level parcel extraction quality, but also strengthens boundary recovery and object-level structural consistency in complex terraced scenes.

The qualitative results in Fig.~\ref{fig:model_compare} further explain the performance differences among methods. Existing approaches mainly suffer from two typical failure modes in complex terraced scenes. The first is \textbf{semantic confusion} caused by the absence of sufficient textual guidance, where visually similar non-parcel regions are easily misclassified as target parcels. The second is \textbf{incomplete structural recovery} caused by the lack of DEM cues, which often leads to discontinuous terrace steps, broken boundaries, and under-segmentation in complex slope regions. By contrast, ETTerra jointly leverages image, text, and DEM information to produce more complete and structurally consistent parcel delineation results.

\begin{figure}[t]
    \centering
    \includegraphics[width=1.0\linewidth]{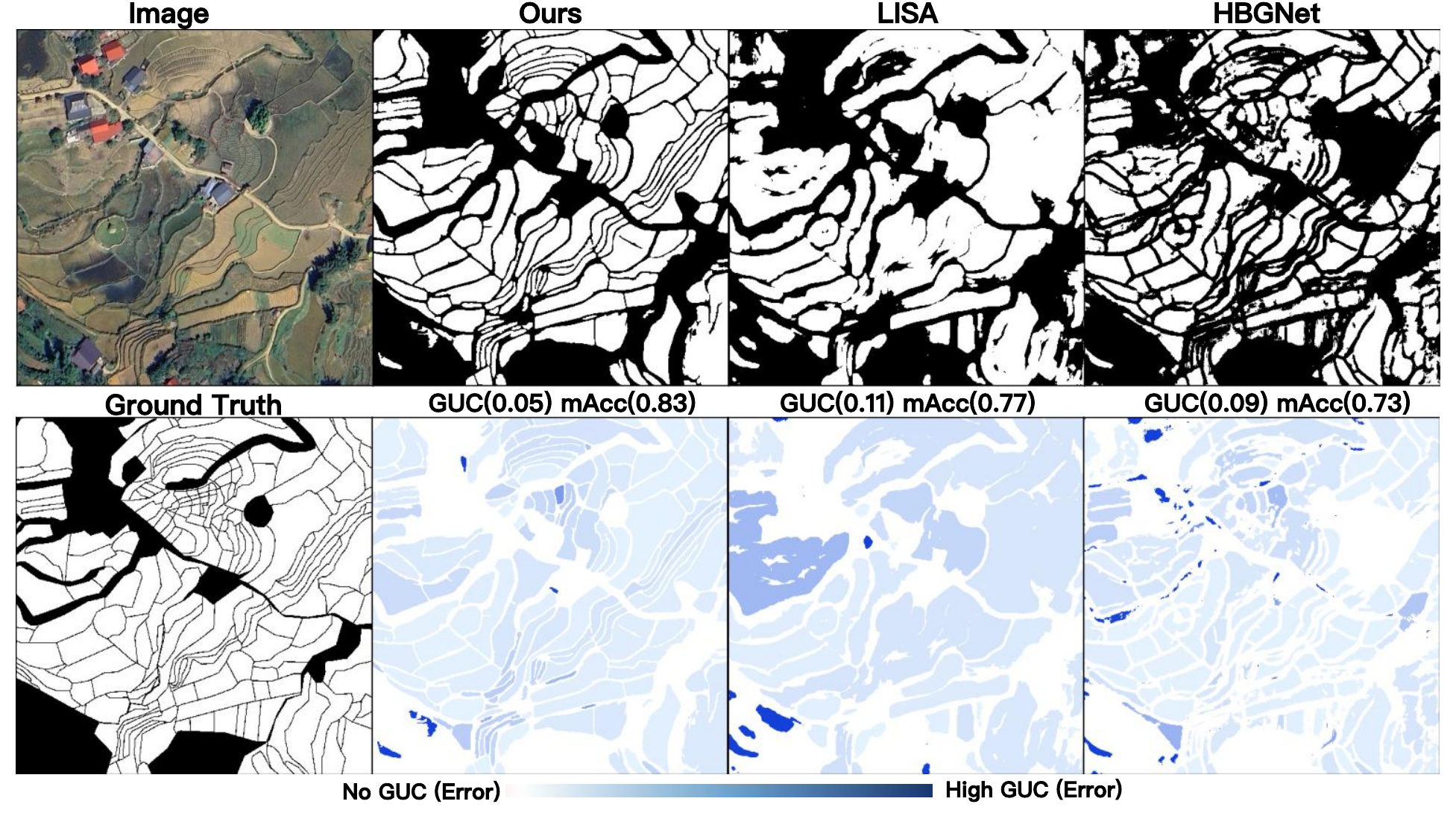}
    \caption{Object-level under-segmentation error analysis of different methods on GTPBD-MM. We visualize GUC-based error distribution together with GUC and mAcc. }
    \vspace{-1.8em}
    \label{fig:object}
\end{figure}

These advantages are more clearly reflected in Fig.~\ref{fig:edge} and Fig.~\ref{fig:object}. While already maintaining the best overall pixel-level performance, ETTerra further recovers more correct edges and significantly reduces both false positive and false negative edges, leading to the best OIS and ODS results. More importantly, the improved boundary recovery also translates into better object-level structure quality, yielding clearer separation between adjacent parcels and better preservation of parcel completeness. As shown in Fig.~\ref{fig:object}, the GUC errors of ETTerra are concentrated in fewer local regions, whereas competing methods are more prone to parcel merging and missing narrow terrace structures, resulting in more severe under-segmentation. In other words, the gain in boundary quality is not limited to edge-level metrics, but also directly contributes to more robust object-level modeling. 

Overall, these results demonstrate the strong complementarity between \textbf{textual semantics} and \textbf{terrain geometry} for complex terraced parcel extraction: the former helps reduce semantic confusion in challenging scenes, while the latter is crucial for fine-grained boundary recovery and object completeness modeling. Built upon this complementarity, ETTerra is able to integrate multimodal cues more effectively and thus achieves consistent advantages in region segmentation, boundary delineation, and object-level modeling. More visualization results could be found in Appendix D.

\section{Conclusion}

We present \textbf{GTPBD-MM}, the first multimodal dataset and benchmark for complex terraced parcel extraction, which jointly aligns high-resolution imagery, structured text descriptions, and DEM data under a unified evaluation framework. GTPBD-MM supports three benchmark settings, namely {Image-only}, {Image+Text}, and {Image+Text+DEM}, enabling systematic analysis of the complementary roles of appearance, semantics, and geometry in terraced parcel extraction. We further propose \textbf{E}levation-\textbf{T}ext guided \textbf{Terra}ced parcel network (\textbf{ETTerra}) as a unified multimodal baseline.
Experimental results show that textual semantics and terrain geometry provide effective complementary cues beyond visual appearance alone, leading to more accurate, coherent, and structurally consistent parcel delineation in complex terraced scenes. We believe GTPBD-MM can serve as a valuable benchmark for future multimodal remote sensing in complex agricultural terrains.

\clearpage
\section{Appendices}
\appendix

\section{More Image Cases}

\begin{figure*}[t]
    \centering
    \includegraphics[width=1.0\linewidth]{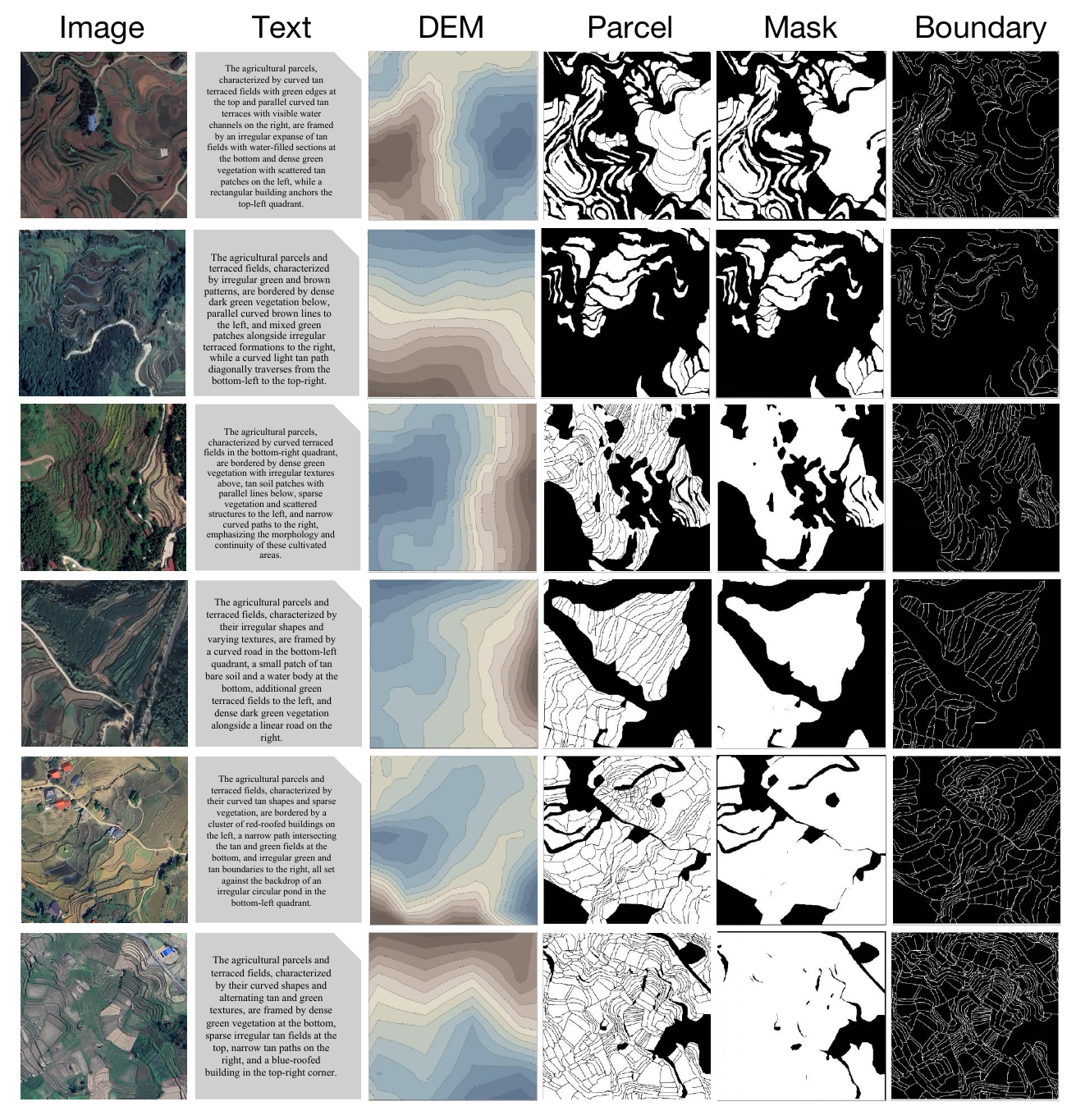}
    \caption{More representative cases of GTPBD-MM from different regions. From top to bottom, the samples are collected from Chongqing, Guangxi, Zhejiang, and Guangdong in China, followed by Vietnam and Indonesia.}
    \label{fig:more_example}
\end{figure*}

Figure~\ref{fig:more_example} presents more representative samples of GTPBD-MM from different regions, including Chongqing, Guangxi, Zhejiang, and Guangdong in China, as well as Vietnam and Indonesia. From left to right, each row shows the optical image, the corresponding task-oriented text description, the aligned DEM, and the three levels of annotations, namely parcel, mask, and boundary. These examples further illustrate the diversity of GTPBD-MM across different terraced landscapes, where parcel morphology, terrain variation, surrounding land-cover patterns, and boundary complexity vary substantially across regions. At the same time, the figure highlights the unified multimodal organization of the dataset, where image appearance, textual semantics, terrain geometry, and hierarchical annotations are spatially and semantically aligned within each sample. Such a design provides a consistent data basis for studying multimodal terraced parcel extraction under diverse geographic and geomorphological conditions.

\begin{figure*}[t]
    \centering
    \includegraphics[width=1\linewidth]{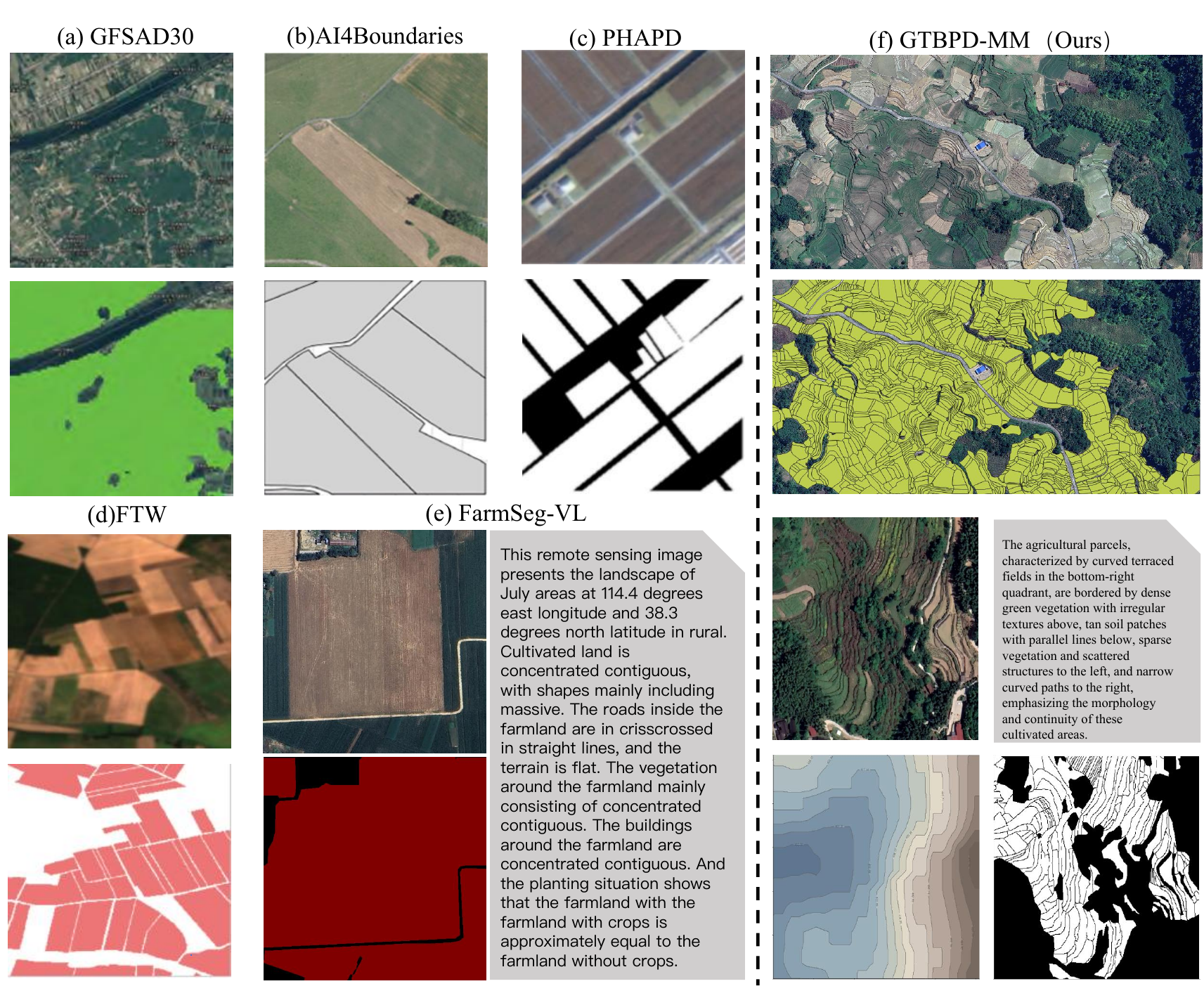}
    \caption{Visual comparison between representative agricultural parcel datasets and GTPBD-MM. Existing datasets mainly focus on regular or relatively flat farmland scenes and provide limited modality or annotation forms, whereas GTPBD-MM presents more complex terraced parcels together with aligned image, text, DEM, and fine-grained annotations.}
    \label{fig:dataset_compare}
\end{figure*}

Figure~\ref{fig:dataset_compare} provides a visual comparison between GTPBD-MM and several representative agricultural parcel datasets. As shown in the figure, existing datasets such as GFSAD30~\cite{GFSAD30}, AI4Boundaries~\cite{AI4Boundaries}, PHAPD\cite{FHAPD/HBGNet}, and FTW\cite{FTW} mainly present regular or relatively flat farmland scenes, while FarmSeg-VL\cite{farmsegvl} further introduces the text modality for farmland understanding. In contrast, GTPBD-MM is designed for more complex terraced landscapes with irregular parcel boundaries, stronger terrain variation, and richer structural patterns. Moreover, GTPBD-MM organizes optical imagery, task-oriented text descriptions, aligned DEM, and fine-grained annotations in a unified sample, which more clearly reflects the multimodal and structurally complex nature of terraced parcel extraction. This comparison further highlights the distinctiveness of GTPBD-MM from previous agricultural benchmarks.

\section{Model details}
\subsection{ETTerra}

ETTerra is implemented under the full \textit{Image+Text+DEM} setting of GTPBD-MM. In our implementation, CLIP~\cite{clip} is adopted as the text encoder to extract language-guided semantic features, while SAM~\cite{sam} is used as the segmentation backbone and mask decoder. Given a spatially aligned RGB image, DEM map, and text description, the semantic branch generates text-guided prompts from cross-modal features, and the DEM branch enhances dense visual features with terrain-aware modulation. The modulated features are further aggregated with the original visual features through a zero-initialized residual connection, and are then jointly fed into the SAM mask decoder together with the text-guided prompts to produce the final terraced parcel mask. Unless otherwise specified, all pretrained backbones are initialized from their official checkpoints. The remaining training and inference settings are summarized in Table~\ref{tab:etterra_details}. 
\begin{table}[t]
\centering
\caption{Hyperparameter settings of ETTerra on GTPBD-MM.}
\label{tab:etterra_details}
\resizebox{\linewidth}{!}{
\begin{tabular}{l l l}
\toprule
\textbf{Category} & \textbf{Setting} & \textbf{Value} \\
\midrule
Input & Image size & $512 \times 512$ \\
Input & DEM size & $512 \times 512$ \\
Input & Text max length & 77 \\
Backbone & Text encoder & CLIP \\
Backbone & Segmentation backbone  & SAM \\
Optimizer & Type & AdamW \\
Optimizer & Learning rate & $1\times10^{-4}$ \\
Optimizer & Weight decay & $1\times10^{-2}$ \\
Training & Batch size & 8 \\
Training & Epochs & 300 \\
Training & LR scheduler & Cosine annealing with linear warm-up \\
Training & Data augmentation & Random horizontal / vertical flipping \\
Loss & Objective & BCE + Dice \\
Inference & Model selection & Last checkpoint on validation set \\
Framework & Implementation & PyTorch \\
Hardware & GPU & NVIDIA RTX 4090 \\
\bottomrule
\end{tabular}
}
\vspace{-1em}
\end{table}

\vspace{-1em}
\subsection{Baseline Methods}
Following the benchmark protocol of GTPBD-MM, we compare eleven baseline methods from four methodological families, including five general semantic segmentation models (U-Net~\cite{unet}, PSPNet\cite{pspnet}, DeepLabV3\cite{deeplabv3}, SegFormer\cite{segformer}, and Mask2Former\cite{mask2former}), two parcel delineation models (REAUNet\cite{reaunet} and HBGNet\cite{FHAPD/HBGNet}), three reasoning segmentation models (LaSagnA\cite{wei2024lasagna}, LISA\cite{lai2024lisa}, and PixelLM\cite{ren2024pixellm}), and one multimodal parcel delineation model (FSVLM\cite{fsvlm/fit}). The first two families are evaluated under the \textit{Image-only} setting, whereas the latter two are evaluated under the \textit{Image+Text} setting.

For the general semantic segmentation baselines, U-Net is adopted as a classical encoder--decoder network with skip connections for recovering fine spatial details. PSPNet\cite{pspnet} is used to aggregate multi-scale contextual priors through pyramid pooling. DeepLabV3\cite{deeplabv3} employs atrous convolution together with ASPP to enlarge the receptive field and improve multi-scale representation. SegFormer\cite{segformer} adopts a hierarchical Transformer encoder with a lightweight MLP decoder, while Mask2Former\cite{mask2former} performs mask prediction with a masked-attention Transformer decoder. For fair comparison, these five image-only baselines are retrained under a unified setting on GTPBD-MM, using $512\times512$ patches, random mirroring and rotation augmentation, SGD with momentum $0.9$ and weight decay $10^{-4}$, on NVIDIA RTX 4090 GPUs.

For parcel delineation, REAUNet\cite{reaunet} and HBGNet\cite{FHAPD/HBGNet} are included as two task-specific baselines. REAUNet\cite{reaunet} is an edge-aware convolutional framework that enhances a U-Net-style backbone with edge detection, dual attention, and refinement modules for agricultural parcel delineation. In our experiments, REAUNet\cite{reaunet} follows the released training configuration with a batch size of 8, a learning rate of $3\times10^{-4}$, weight decay of $10^{-4}$, a step-based decay schedule with $\gamma=0.1$, and a maximum of 200 epochs. HBGNet\cite{FHAPD/HBGNet} is a hierarchical semantic boundary-guided network with a parcel branch and an auxiliary boundary branch, and further introduces a Laplacian-based boundary extraction mechanism together with a PVT-v2 backbone. We follow its released setting with Adam optimizer, a learning rate of $10^{-4}$, batch size 8, 100 training epochs, and a cosine annealing scheduler with $\eta_{\min}=10^{-5}$.

For reasoning segmentation, we use LaSagnA\cite{wei2024lasagna}, LISA\cite{lai2024lisa}, and PixelLM\cite{ren2024pixellm}. LaSagnA\cite{wei2024lasagna} is a language-based segmentation assistant designed for complex language queries. Its released implementation builds upon LLaVA-7B and SAM-ViT-H, fine-tunes the language model with LoRA~\cite{hu2022lora}, and trains the SAM\cite{sam} decoder jointly; the public training script uses DeepSpeed with batch size 2 and model maximum length 1024, while the remaining optimization details follow the released codebase. LISA\cite{lai2024lisa} extends multimodal large language models to reasoning segmentation by introducing a segmentation token for mask prediction. Following its official paper, we use the LLaVA7B setting with a SAM ViT-H backbone, fine-tune the model with a learning rate of $3\times10^{-4}$ and zero weight decay, use WarmupDecayLR with 100 warm-up iterations, batch size 2 per device with gradient accumulation of 10, and optimize the text, mask, BCE, and Dice losses jointly. PixelLM\cite{ren2024pixellm} further improves pixel-level reasoning by combining a CLIP-ViT-L/14 vision encoder, a multimodal LLM, a segmentation codebook, and a lightweight pixel decoder. We follow its released configuration using AdamW, learning rate $3\times10^{-4}$, zero weight decay, betas $(0.9,0.95)$, batch size 16, WarmupDecayLR with 100 warm-up steps, and no additional data augmentation.

For multimodal parcel delineation, we include FSVLM\cite{fsvlm/fit} as an image--text baseline for farmland segmentation. FSVLM\cite{fsvlm/fit} combines multimodal language modeling with segmentation-oriented remote sensing parsing, and its released implementation uses LLaVA-7B, a CLIP\cite{clip} vision tower, and SAM-ViT-H initialization. In our experiments, we adapt FSVLM\cite{fsvlm/fit} to the unified benchmark setting of GTPBD-MM by using an input image size of $512\times512$, while the text input is encoded according to the text organization of our dataset. The remaining optimization settings generally follow its released training configuration.

\section{Evaluation metrics}
\label{app:evaluate}

\subsection{Pixel--level evaluation metrics}
Following the benchmark protocol in the main paper, we adopt five pixel-level metrics for evaluating region-level segmentation quality on GTPBD-MM, including Recall \textbf{(Rec.)}, \textbf{F1-score}, Overall Accuracy \textbf{(OA)}, mean Intersection over Union \textbf{(mIoU)}, and mean Accuracy \textbf{(mAcc)}. These metrics provide complementary views of pixel-wise segmentation performance from the perspectives of completeness, overlap quality, and class-balanced accuracy. In our binary setting, the two classes correspond to parcel and non-parcel regions.

\textbf{Recall (Rec.)} measures the proportion of true parcel pixels that are correctly identified:
\begin{equation}
    Rec. = \frac{TP}{TP + FN},
\end{equation}
where $TP$ and $FN$ denote true positive and false negative, respectively.

\textbf{F1-score} is the harmonic mean of precision and recall, and provides a balanced evaluation of pixel-wise prediction quality:
\begin{equation}
    F1 = \frac{2TP}{2TP + FP + FN},
\end{equation}
where $FP$ denotes false positive.

\textbf{Overall Accuracy (OA)} calculates the proportion of correctly classified pixels over the entire image:
\begin{equation}
    OA = \frac{TP + TN}{TP + FP + FN + TN},
\end{equation}
where $TN$ denotes true negative.

\textbf{mean Intersection over Union (mIoU)} evaluates the average overlap quality across classes:
\begin{equation}
    mIoU = \frac{1}{C} \sum_{c=1}^{C} \frac{TP_c}{TP_c + FP_c + FN_c},
\end{equation}
where $C$ is the number of classes, and $TP_c$, $FP_c$, and $FN_c$ denote the true positive, false positive, and false negative pixels of class $c$, respectively.

\textbf{mean Accuracy (mAcc)} measures the average per-class recall:
\begin{equation}
    mAcc = \frac{1}{C} \sum_{c=1}^{C} \frac{TP_c}{TP_c + FN_c}.
\end{equation}

\subsection{Edge--level evaluation metrics}
For edge detection tasks on GTPBD-MM, we evaluate model performance using two widely adopted metrics: Optimal Dataset Scale F1-score (ODS) and Optimal Image Scale F1-score (OIS). These metrics assess the quality of predicted boundaries from both dataset-level and image-level perspectives.

\noindent
Let $P_t$ and $R_t$ denote precision and recall computed at threshold $t$, and let $F_t$ be the corresponding F1-score:
\begin{equation}
F_t = \frac{2 \cdot P_t \cdot R_t}{P_t + R_t}.
\end{equation}

\noindent
\textbf{Optimal Dataset Scale F1-score (ODS)} evaluates the best dataset-level boundary performance under a single threshold:
\begin{equation}
\text{ODS} = \max_{t \in \mathcal{T}} \left( \frac{2 \cdot P_t^{\text{dataset}} \cdot R_t^{\text{dataset}}}{P_t^{\text{dataset}} + R_t^{\text{dataset}}} \right),
\end{equation}
where $P_t^{\text{dataset}}$ and $R_t^{\text{dataset}}$ are the aggregated precision and recall over the whole dataset at threshold $t$.

\noindent
\textbf{Optimal Image Scale F1-score (OIS)} computes the average of the best per-image F1-scores:
\begin{equation}
\text{OIS} = \frac{1}{N} \sum_{i=1}^{N} \max_{t \in \mathcal{T}} \left( \frac{2 \cdot P_t^{(i)} \cdot R_t^{(i)}}{P_t^{(i)} + R_t^{(i)}} \right),
\end{equation}
where $P_t^{(i)}$ and $R_t^{(i)}$ denote the precision and recall of the $i$-th image under threshold $t$, and $N$ is the total number of images.

\subsection{Object--level geometric metrics}
To evaluate the geometric quality of delineated terraced parcels, we adopt three object-level metrics: Global Over-Classification Error (GOC), Global Under-Classification Error (GUC), and Global Total Classification Error (GTC). These metrics quantify object-level errors in terms of spatial overreach, omission, and overall structural inconsistency.

Let $S_i$ denote the $i$-th predicted parcel and let $O_i$ denote the ground-truth parcel with the largest overlap with $S_i$. Let $m$ be the number of predicted parcels.

\noindent
\textbf{Global Over-Classification Error (GOC)} measures the extent to which a predicted parcel exceeds its matched ground-truth object:
\begin{equation}
\text{OC}(S_i) = 1 - \frac{\text{area}(S_i \cap O_i)}{\text{area}(S_i)},
\end{equation}
\begin{equation}
\text{GOC} = \sum_{i=1}^{m} \left( \text{OC}(S_i) \cdot \frac{\text{area}(S_i)}{\sum_{k=1}^{m} \text{area}(S_k)} \right),
\end{equation}
where $\text{area}(\cdot)$ denotes the number of pixels in the corresponding region.

\noindent
\textbf{Global Under-Classification Error (GUC)} measures the extent to which the matched ground-truth parcel is not fully covered by the prediction:
\begin{equation}
\text{UC}(S_i) = 1 - \frac{\text{area}(S_i \cap O_i)}{\text{area}(O_i)},
\end{equation}
\begin{equation}
\text{GUC} = \sum_{i=1}^{m} \left( \text{UC}(S_i) \cdot \frac{\text{area}(S_i)}{\sum_{k=1}^{m} \text{area}(S_k)} \right).
\end{equation}

\noindent
\textbf{Global Total Classification Error (GTC)} combines over-classification and under-classification errors into a unified metric:
\begin{equation}
\text{TC}(S_i) = \sqrt{\frac{\text{OC}(S_i)^2 + \text{UC}(S_i)^2}{2}},
\end{equation}
\begin{equation}
\text{GTC} = \sum_{i=1}^{m} \left( \text{TC}(S_i) \cdot \frac{\text{area}(S_i)}{\sum_{k=1}^{m} \text{area}(S_k)} \right).
\end{equation}

\section{More Results}

\subsection{More Boundary Visualization Results}
\begin{figure*}
    \centering
    \includegraphics[width=1\linewidth]{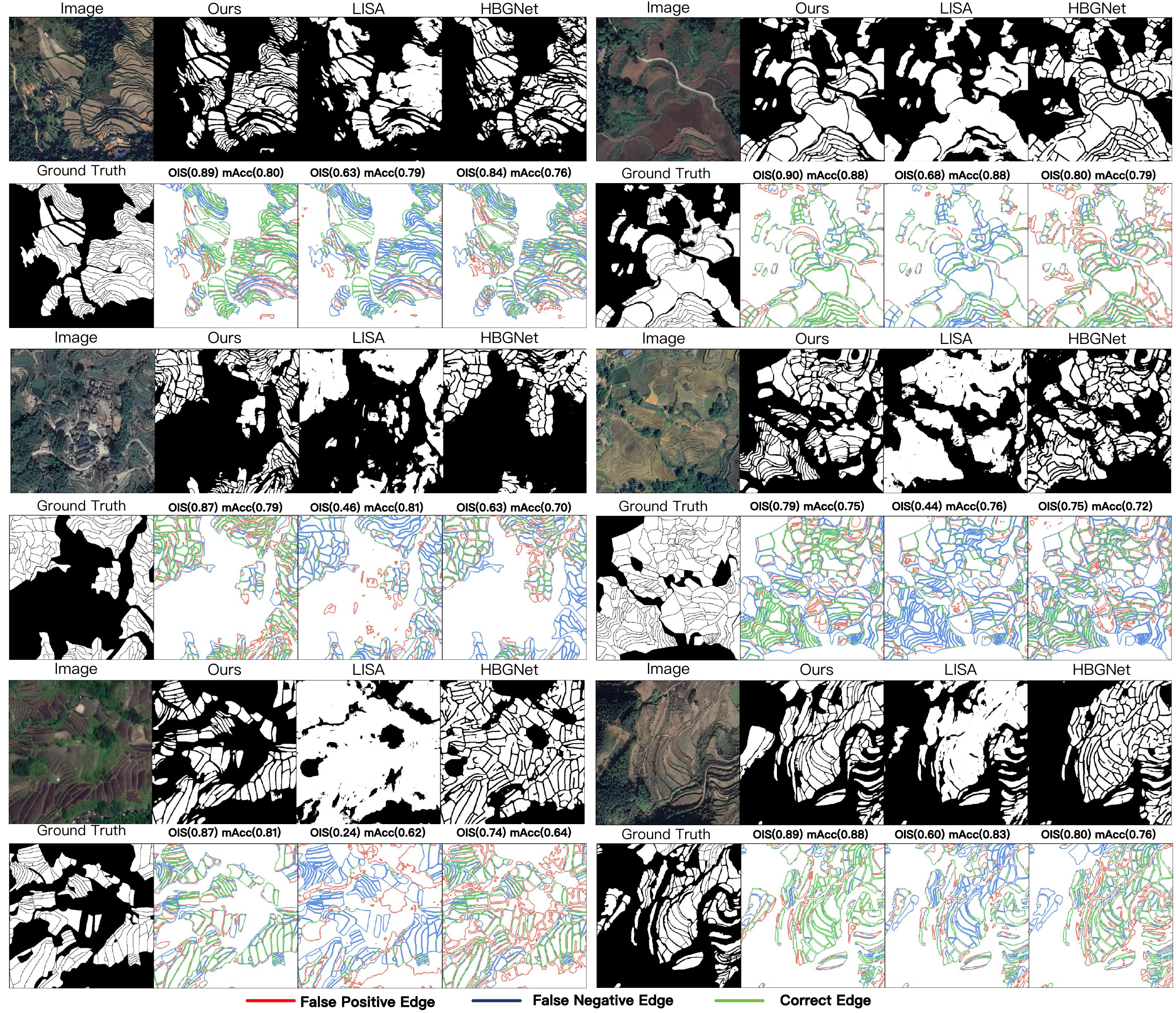}
    \caption{More boundary visualization results on representative regions. For each case, we compare the predictions of Ours, LISA, and HBGNet with the ground truth. The bottom row in each case shows edge-level error visualization, where red, blue, and green denote false positive edges, false negative edges, and correct edges, respectively.}
    \label{fig:edgemore}
    \vspace{-1em}
\end{figure*}
Figure~\ref{fig:edgemore} presents more qualitative comparisons of boundary visualization results on representative regions. For each case, we compare our method with LISA and HBGNet, together with the corresponding ground-truth parcel mask and edge map. To better analyze the quality of boundary delineation, we further visualize the edge-level errors using color coding, where red denotes false positive edges, blue denotes false negative edges, and green denotes correctly predicted edges. As shown in Fig.~\ref{fig:edgemore}, our method generally produces more complete and structurally consistent parcel boundaries, while reducing both missing terrace edges and redundant boundary responses in complex regions. In contrast, LISA tends to miss fine-grained terrace boundaries, and HBGNet, although stronger in boundary awareness, still suffers from fragmented or locally inaccurate delineation in highly irregular scenes.

\subsection{More Object-level Error Visualization Results}
\begin{figure*}
    \centering
    \includegraphics[width=1\linewidth]{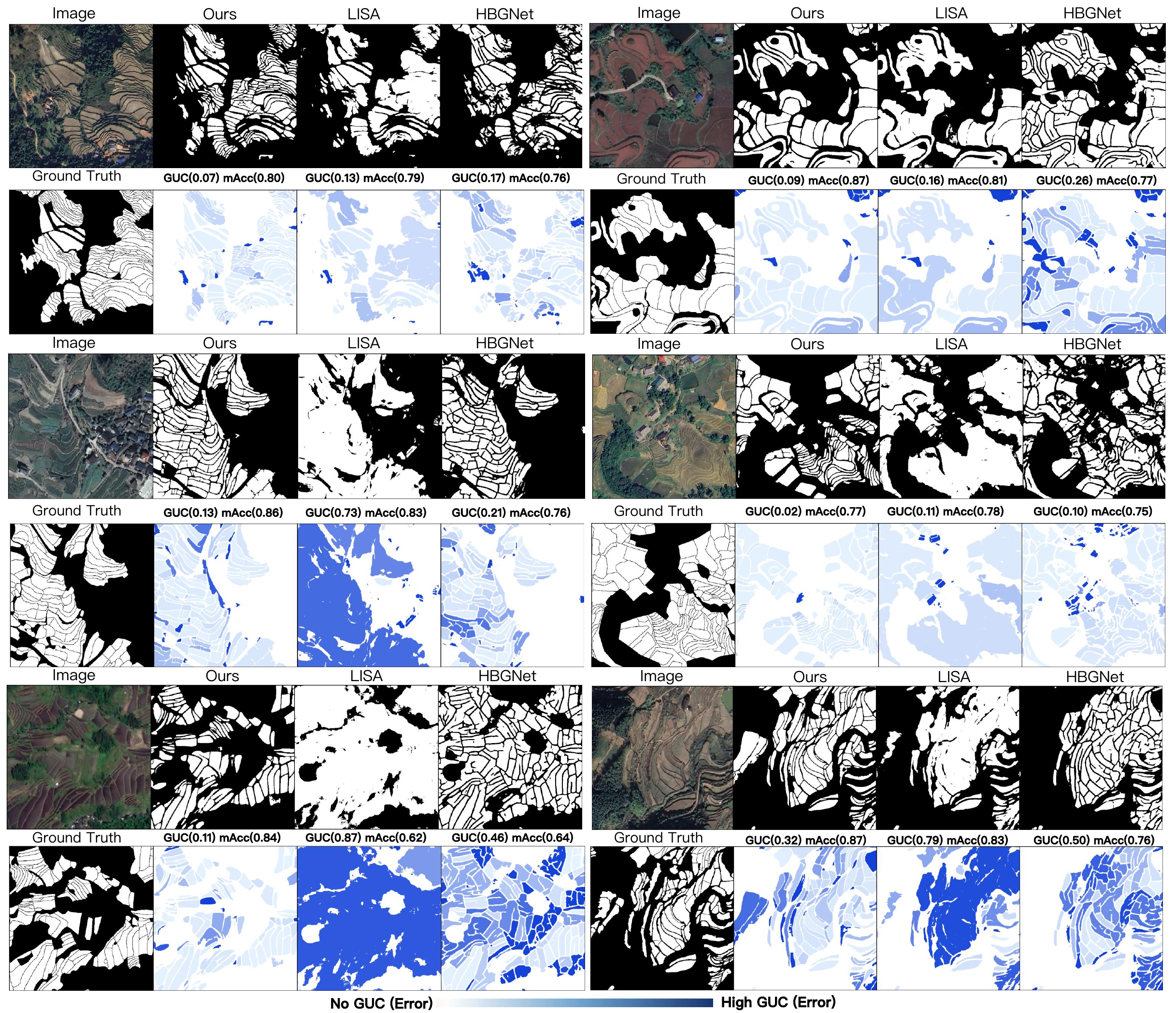}
    \caption{More object-level error visualization results on representative regions. For each case, we compare the predictions of Ours, LISA, and HBGNet with the ground truth. The error maps are colored according to GUC values, where darker blue indicates larger under-segmentation errors.}
    \label{fig:objectmore}
    \vspace{-1em}
\end{figure*}
Figure~\ref{fig:objectmore} shows more object-level error visualizations on representative regions. We compare Ours, LISA, and HBGNet from the perspective of geometric consistency at the parcel level. Specifically, we use GUC-based error maps to highlight under-segmentation regions, where darker blue indicates more severe object-level errors. As illustrated in Fig.~\ref{fig:objectmore}, our method usually yields lower object-level errors and preserves parcel completeness more effectively, especially in terraced scenes with curved structures, adjacent parcels, and complex topological layouts. By contrast, the compared methods more easily suffer from parcel merging or incomplete delineation, which leads to larger under-segmentation errors in challenging regions.

\section{Limitations and Future Work}

The current version of GTPBD-MM mainly focuses on establishing a unified benchmark for multimodal terraced parcel extraction under the image--text--DEM setting, and ETTerra is designed as a benchmark baseline to verify the effectiveness of jointly modeling appearance, semantics, and terrain geometry. While this setting is sufficient for systematic evaluation in the present study, there is still room to further extend both the benchmark and the modeling framework in broader directions.

In future work, we plan to further enrich GTPBD-MM from both the data and model perspectives. On the data side, a natural direction is to expand the benchmark to more regions, more diverse terraced styles, and more complex scene conditions, so as to support broader cross-region analysis. We also plan to enrich the text modality with more diverse and fine-grained descriptions, enabling stronger multimodal understanding and reasoning beyond the current task-oriented setting. In addition, incorporating temporal observations or additional auxiliary modalities may provide a more comprehensive foundation for terraced scene understanding.

On the model side, future research may explore stronger multimodal foundation models and more advanced fusion strategies for jointly leveraging image, text, and terrain information. Another promising direction is to move beyond raster-level delineation toward topology-aware, boundary-preserving, or vectorization-oriented parcel extraction. We also expect this benchmark to facilitate future studies on cross-region generalization, domain adaptation, weakly supervised learning, and multimodal agricultural geospatial intelligence.

\clearpage
\bibliographystyle{ACM-Reference-Format}
\bibliography{main}










\end{document}